\def\year{2024}\relax
\documentclass[letterpaper]{article} 
\usepackage{aaai24}  
\usepackage{times}  
\usepackage{helvet}  
\usepackage{courier}  
\usepackage[hyphens]{url}  
\usepackage{graphicx} 
\usepackage{natbib}  
\usepackage{caption} 
\DeclareCaptionStyle{ruled}{labelfont=normalfont,labelsep=colon,strut=off} 
\usepackage{algorithm}
\usepackage{algorithmic}

\usepackage{newfloat}
\usepackage{listings}

\usepackage[inkscapeexe=/Applications/Inkscape.app/Contents/MacOS/inkscape,inkscapearea=page]{svg}

\floatstyle{ruled}
\newfloat{listing}{tb}{lst}{}
\floatname{listing}{Listing}

\usepackage[inkscapearea=page]{svg}
\usepackage{adjustbox}
\usepackage{relsize}
\usepackage{subcaption}
\usepackage{makecell}
\usepackage{comment}

\nocopyright

\title{
    Search-based versus Sampling-based Robot Motion Planning:\\ 
    A Comparative Study
}
\author{
    Georgios Sotirchos\textsuperscript{\rm 1},
    Zlatan Ajanovi\'c \textsuperscript{\rm 2} \\
}
\affiliations{
    \textsuperscript{\rm 1}Delft University of Technology \\
    \textsuperscript{\rm 2}RWTH Aachen University \\
    g.sotirchos@student.tudelft.nl, zlatan.ajanovic@ml.rwth-aachen.de
}

\definecolor{MyOrange}{rgb}{1.0, 0.5, 0.0}

\begin{document}

\maketitle

\begin{abstract}

\begin{quote}

Robot motion planning is a challenging domain as it involves dealing with high-dimensional and continuous search space. In past decades, a wide variety of planning algorithms have been developed to tackle this problem, sometimes in isolation without comparing to each other. In this study, we benchmark two such prominent types of algorithms: OMPL’s sampling-based RRT-Connect and SMPL’s search-based ARA* with motion primitives. To compare these two fundamentally different approaches fairly, we adapt them to ensure the same planning conditions and benchmark them on the same set of planning scenarios. Our findings suggest that sampling-based planners like RRT-Connect show more consistent performance across the board in high-dimensional spaces, whereas search-based planners like ARA* have the capacity to perform significantly better when used with a suitable action-space sampling scheme. Through this study, we hope to showcase the effort required to properly benchmark motion planners from different paradigms thereby contributing to a more nuanced understanding of their capabilities and limitations. The code is available at \url{https://github.com/gsotirchos/benchmarking_planners}

\end{quote}

\end{abstract}

Motion planning for robotic manipulators involves determining a feasible 
path from a starting configuration to a goal configuration while avoiding 
collisions with obstacles and obeying kinematic and dynamic 
constraints. This problem is particularly challenging due to the 
high-dimensional and continuous configuration space of manipulators, complex 
geometrical environments, and the real-time requirements on computation time. The 
complexity of the problem is further amplified when considering the need for 
discretization of the configuration space, and the cases with dynamic environments where obstacles and goals may change over time, requiring real-time re-planning capabilities.

Modern motion planning algorithms have been often categorized into 
two main families: search-based and sampling-based methods \citep{lavalle_planning_2006}. This categorization might be misleading because both of them search and sample the continuous configuration space. Search-based methods, discretize the space and search for a feasible path in a graph, using algorithms such as Dijkstra's \citep{dijkstra1959note}, A* \citep{hart1968formal}, and their variants. These methods are complete and optimal in the discretized space but suffer from the ``curse of dimensionality'' as the size of the discretized space grows exponentially with the number of dimensions. However, different variants of search-based planning were successfully used for planning footsteps for humanoid robots \citep{ranganeni2020effective},
robot manipulation \citep{mandalika2018lazy}, underwater vehicles \citep{youakim2020multirepresentation}, 
the aggressive flight of UAVs \citep{liu2018search},
as well as for automated driving in unstructured environment \cite{dolgov2008practical, likhachev2009planning, adabala2023multi}, urban environments \citep{ajanovic2018search} and performance driving including drifting maneuvers \citep{ajanovic2019search, ajanovic2023search}.

Sampling-based methods, on the other hand, randomly sample possible configurations
and build a graph that approximates the connectivity 
of the space. Popular algorithms in this family include the Probabilistic 
Roadmap Method (PRM) \citep{kavraki_probabilistic_1996} and Rapidly-exploring Random Trees (RRT) \citep{lavalle_rapidly-exploring_1998}. These 
methods are particularly suited for high-dimensional spaces as they avoid 
explicit discretization. While they are not complete, they are 
probabilistically complete, meaning that they can find a solution with 
probability approaching one as the number of samples increases. Moreover, variants like RRT* and PRM*  \citep{karaman2010incremental} can provide asymptotically optimal paths. Informed variants have been proposed, which improve RRT*'s convergence rate and final solution quality by focusing the search through directed sampling in a subset, such as an ellipsoidal defined based on previous solutions \citep{gammell2014informed}. Particularly useful in robotics are real-time variants of RRT*, in which tree expansion and action-taking are interleaved \citep{naderi_rt-rrt_2015}.


Both families of algorithms have their strengths and 
applications. However, there is fairly limited work towards experimentally 
comparing the two families in a systematic manner. These algorithms are 
developed and evaluated using different software frameworks by different 
research groups and the experimental results of each method are often not 
directly comparable to those of the other.

The main contributions of our work are threefold:

\begin{itemize}
    \item We present a generic categorization of existing planning methods based on their functional architecture and relation of search and sampling steps.
    \item We demonstrate the process and the challenges for objectively comparing the performance of two planning algorithms developed in fundamentally different frameworks.
    \item We provide a fair experimental comparison between the search-based Anytime Repairing A* (ARA*) and the sampling-based RRT-Connect.
\end{itemize}

We organize the paper as follows. In the first and second sections, we outline the main components of the 
robot motion planning problem. In the second section we describe the basic 
components present in every robot motion planning algorithm along with 
a generic representation. Based on this, we categorize existing planning algorithms into four main types and explain their distinguishing 
characteristics. In the third section, we present our benchmarking 
method with the implementation details and the structure of our experiments are recorded. Next, the results of the 
experiments are shown and discussed in the fifth section, and in the sixth section related work to ours is analyzed. Lastly, closing remarks and possible directions of future work are presented in the seventh section.

\section{Robot Motion Planning Problem}

A robot comprises links and joints that can occupy various configurations, or states, depending on the position or velocity of their joints. The collection of all possible states of a system is known as the state space, denoted as $X$, with individual states represented as $x$. Not all states within the state space are physically feasible, as some may violate spatial or otherwise specified constraints. These constraints divide the state space $X$ into the constraint-free region $X_\mathrm{free}$ and its complement $X \setminus X_\mathrm{free}$.

A motion planning problem is defined as a tuple $(X_\mathrm{free}, x_\mathrm{I}, X_\mathrm{G})$, which represents the task of finding a path, a continuous function $p : [0, 1] \rightarrow X_\mathrm{free}$, from a start state $x_\mathrm{I} \in X_\mathrm{free}$ to a goal region $X_\mathrm{G} \subseteq X_\mathrm{free}$ \citep{lavalle_planning_2006}. The set of all feasible paths $P$ is referred to as the path space $P(X_\mathrm{free}, x_\mathrm{I}, X_\mathrm{G})$.

There are several variations of motion planning problems, including:
\textit{i}) Path planning: This variation addresses the geometric problem, ignoring the system's velocity, time, or dynamics, and is often called the piano mover’s problem.
\textit{ii}) Kinodynamic planning: This involves planning for a dynamical system with potential constraints on velocity, acceleration, or torque. A kinodynamic planning problem is defined as a tuple $(X_\mathrm{free}, x_\mathrm{I}, X_\mathrm{G}, f)$, where $f$ represents the dynamical equations.
\textit{iii}) Optimal planning: This focuses on finding a globally optimal path. An optimal path is one that minimizes a given cost functional $c : P \rightarrow R \geq 0$. An optimal motion planning problem is represented as a tuple $(X_\mathrm{free}, x_\mathrm{I}, X_\mathrm{G}, c)$, where the objective is to find a feasible path $p^*$ such that $c(p^*) = c^*$, with $c^*$ being the minimum cost across the path space $P$.

\subsection{Robot Models}

A robot model is an abstraction, capturing both the physical form and the 
functional attributes of a robot, including its geometric configuration and 
kinematic chain. It is required for accurately simulating how the robot can move and
interact with the surrounding environment, thereby enabling the planning of 
paths while respecting the robot's and environment constraints. Central to robot modeling 
is the notion of configuration space (C-space) as defined by \citet{cox_spatial_1990}, which represents every 
possible configuration $q$ the robot might assume across 
a multidimensional space, each dimension of which corresponds to a specific 
degree of freedom (DOF) of the robot, enabling a holistic representation of 
its potential movements. In essence, the robot model is instrumental in the 
collision detection process, assisting in the clear distinction between 
navigable areas and those restricted either by obstacles or by the robot’s 
joint limits. This delineation is essential for obtaining paths conforming to the planning scenario’s requirements.

\subsection{Environment models}

Environment models in robot motion planning are representations of the 
world in which the robot operates, including the obstacles, terrain, and 
other elements that might affect the robot's movement. These models are 
essential for a robot to determine a path from its starting location to its 
destination without colliding with any obstacles. In practice, they are used 
by motion planners in queries for collision checking of new state 
configurations.

In motion planning, the world, denoted as $\mathcal{W}$, includes both the robot and 
the obstacles within it. The space can be two-dimensional for simpler, 
planar motion planning, or three-dimensional for more complex environments 
that require navigation in 3D space, such as in the case of 
manipulators. The obstacle region, represented as $\mathcal{O} \subset \mathcal{W}$, is the 
area within the world that the robot must avoid. This region typically has 
a boundary defined by simple geometric shapes (e.g., polygons in 2D or 
polyhedra in 3D).

Objects contained in the environment can be represented in several ways, 
including analytical geometric descriptions, polygon meshes, or annotated 
voxel spaces.

\subsection{Collision checking}

Collision checking in motion planning is the process that involves determining whether for a given path or motion any link of the robot intersects with another link or obstacles within its environment. Collision checking typically involves evaluating a series of positions or configurations that the robot will occupy along its proposed path. Each of these configurations is then tested to see if the robot would intersect with any known obstacles. This can be computationally intensive, as it may require checking numerous points along the path, especially in complex or dynamically changing environments with a high density of 
obstacles, or in scenarios requiring high precision. Various methods for collision checking exist, ranging from geometric checks in simulation \citep{pan_fcl_2012, hudson_v-collide_1997}, to distance routines like \citet{gilbert_fast_1988} for convex representations, to more complex ones that involve constructing and querying the C-space using learned methods, such as in \citep{benka_direct_2023, danielczuk_object_2021}. The effectiveness and efficiency of collision checking can greatly impact the overall performance of the motion planning system, making it a critical component in the design of autonomous robots.

\section{Motion Planning Algorithms}

A foundational concept in almost all types of planning algorithms are the 
tree data structures due to their utility in solving complex problems by 
breaking down the searched space into manageable segments for exploration 
and decision-making according to \cite{lavalle_planning_2006}. Trees are hierarchical structures composed of nodes and edges, without cycles (as opposed to graphs in general), where each node corresponds to a state 
or a configuration in the space the planner is exploring, and the edges 
represent transitions between them. These trees are constructed and used 
during the planning process, with their root typically representing the 
starting state, and their branches leading to different possible future 
states based on the actions taken. In addition, they allow for 
backtracking, pruning of infeasible paths, and incorporating various 
optimization criteria, as explained by \cite{orthey_sampling-based_2023}.

The state tree can be expanded by obtaining new state configurations. The 
continuous space or domain can be probed by one of several \textbf{sampling} methods on the spectrum between orderly sampling in classical grid search and random sampling as in RRT and PRM, as 
explained by \cite{lavalle_relationship_2004}. This process is particularly relevant in 
motion planning for manipulators, where it is impractical to evaluate 
analytically every possible robot configuration. As discussed by \cite{Kavraki2016}, by intelligently sampling 
the C-space, algorithms can search for feasible paths to navigate, while 
significantly reducing computational load or balancing exploration and 
exploitation in unknown environments.
To construct a viable route, the set of obtained states (whether by 
sampling or discretization) must eventually be subjected to some form of 
\textbf{search}. This search process involves systematically exploring the set of 
states, evaluating the feasibility of transitions, and then incrementally 
building a sequence that satisfies the motion planning criteria. It can be 
guided by various heuristics or strategies to improve efficiency, reduce 
computation time, and increase the likelihood of finding a valid path if 
one exists.

When solving a planning problem, there are several universal elements that 
are utilized by every type of planning algorithm. These are the following:

\begin{itemize}
    \item \textit{Query}: the initial and goal configurations of the robot, 
        $q_{\mbox{\tiny I}}$ and $q_{\mbox{\tiny G}}$\footnote{The goal can also be represented by a set of states $q_{\mbox{\tiny G}} \in G$} respectively, for which the algorithm is requested 
        to plan a path connecting them.
    \item \textit{Path}: the path $p^*$ of successive configurations or 
        actions generated by the planning algorithm.
    \item \textit{Graph}: a graph $G$ with robot configurations (states) as 
        its nodes $V$, and the transitions connecting them as its edges 
        $E$. It is used and updated by the algorithm to during the planning 
        process.
    \item \textit{Geometric models}: representations of the actual 
        objects present in the environment as well as of the robot 
        itself.
    \item \textit{Collision detection}: a method with which the planner can 
        evaluate whether the robot's configuration in a node is colliding 
        with the environment or with itself, based on their 
        representations.
    \item \textit{Sampling}: the process with which the algorithm can 
        obtain new sets of nodes that can be subsequently evaluated, added 
        to the graph, and included in possible paths. The sampling domain 
        is specified by the algorithm and can be the state or the action 
        space.
    \item \textit{Heuristic function}: a function that can provide an 
        estimation of the value of a node that represents some measure of 
        proximity to the goal node.
    \item \textit{Discrete search}: a way to traverse the graph and obtain  
        a sequence of nodes or edges connecting the initial and goal nodes, 
        while optimizing specific metrics (distance, duration, etc.).
    \item \textit{Algorithm}: the logic performed by the planning algorithm 
        itself, arbitrating each of the other elements to produce a path 
        given a planning query.
\end{itemize}

\begin{figure}[ht]
    \centering
    \includesvg[%
        width=\columnwidth,%
        pretex=\relscale{0.85}%
    ]{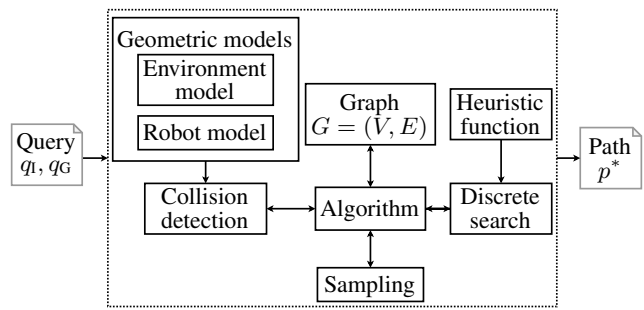}
    \caption{Generic elements of robot motion planning algorithms.}
    \label{fig:generic_planner_diagram}
\end{figure}

Depending on the sequence and the method in which the above processes are performed the different existing robot motion planning algorithms can be classified in four main categories.

\subsection{Sample/Decompose-then-Search Motion Planning}

The basic principle of these methods is the separation of the graph construction phase and the graph search. The two most prominent methods are Voronoi diagram planning and Probabilistic Roadmap Method (PRM).

In Voronoi-based motion planning a Voronoi diagram is constructed based on 
the complete C-space. As formulated by \cite{leven_planning_1987}, the configuration space is partitioned into cells by the 
Voronoi diagram, where each cell contains points that are closer to 
a particular obstacle (or set of obstacles) than to any other. The Voronoi 
diagram essentially divides the space into regions based on the nearest 
neighbor relation with respect to the obstacles. This generates a graph 
where the edges represent paths equidistant from the nearest obstacles, 
ensuring that the path maximizes the clearance from obstacles, which is often 
preferable for safety and maneuverability \citep{bhattacharya_roadmap-based_2008}.

Subsequently, the shortest or most efficient path along these edges, 
connecting the start to the goal position, can be searched for using graph 
search algorithms such as Dijkstra's algorithm or the A* algorithm. This 
approach is particularly useful in environments where navigating close to 
obstacles poses a high risk or is undesirable. However, constructing 
a Voronoi diagram, especially in high-dimensional spaces or complex 
environments, can be computationally intensive. Furthermore, the paths 
generated by Voronoi decomposition may not always be the most direct paths 
between two points, as the primary objective is to maximize clearance from 
obstacles rather than minimize path length.

The Probabilistic Roadmap Method (PRM) is specifically designed to handle 
multiple-query situations efficiently by performing substantial 
preprocessing to build a roadmap. Formulated by 
\cite{kavraki_probabilistic_1996}, this roadmap is an undirected graph 
constructed during a preprocessing phase, where vertices represent random 
configurations in the free workspace, and edges represent collision-free paths 
between these configurations. Once built, this graph can be used 
to solve various subsequent path planning queries within 
the same environment by employing a search algorithm after connecting 
start and goal configurations to the roadmap.

\begin{figure}[ht]
    \centering
    \includesvg[%
        width=\columnwidth,%
        pretex=\relscale{0.85}%
    ]{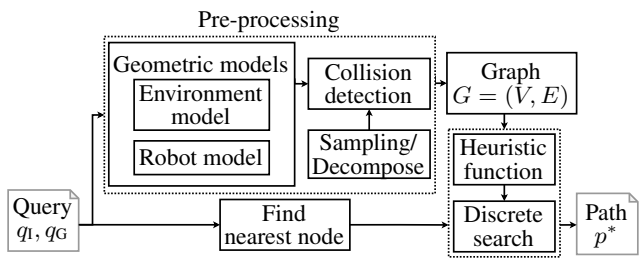}
    \caption{Planning elements of Sample/Decompose-then-Search algorithms.}
    \label{fig:prm_diagram}
\end{figure}

The planning components utilized by Sample/Decompose-then-Search algorithms
are shown in Figure \ref{fig:prm_diagram}. During Voronoi decomposition, the workspace is divided in a structured way, while collision detection is implicit and the geometric models are used directly for obtaining the nodes. It is worth noting that the decomposition process is analogous to sampling as the nodes of the graph are obtained in this phase, albeit in a more systematic manner. By contrast, during PRM's pre-processing phase the whole of 
C-space is sampled randomly and the nodes that are not in collision are 
then added to the graph. Subsequently, each query's start and goal nodes are added to the produced graph which is subjected to the search.

\subsection{Sampling-guided Motion Planning}
The basic principle of these methods is that the sampling dominates the process and practically guides the search. The most prominent algorithm is Rapidly-exploring Random Tree (RRT).
Rapidly exploring Random Tree (RRT) solves the problem of navigating 
a space by incrementally building a tree that explores the available space, 
starting from an initial point and expanding towards unexplored areas. This 
approach, formulated by \cite{lavalle_rapidly-exploring_1998}, is particularly 
well-suited for high-dimensional spaces or environments with complex obstacles. 
We classified RRT (and the algorithms based on it) as purely sampling-based because it constructs a path through the space by randomly sampling points and incrementally building a path or tree towards these points, without any explicit searching steps. This method contrasts with discrete search-based algorithms, which might systematically explore the space or with deterministic methods that might solve a mathematical formulation of the problem directly. By sampling, RRT can efficiently explore large spaces and find paths that avoid obstacles without exhaustive search.

One of the key features of RRT is its Voronoi bias, which refers to the 
tendency of the algorithm to preferentially expand into the largest 
unsearched areas of the space. New nodes are added by randomly sampling the 
obstacle-free space and then connected, to the closest node in the tree provided that the path between them is collision-free effectively 
guiding the tree expansion towards unexplored areas, as discussed by 
\cite{lindemann_incrementally_2004}. The nature of this process means that 
regions of the space that are less densely covered by the tree (i.e., 
parts of the Voronoi diagram with larger cells) are more likely to be 
selected for expansion, ensuring more efficient coverage of the 
space. Thus, it encourages rapid exploration of unvisited areas, helping 
the algorithm to find a path to the goal more quickly than it might 
otherwise and to navigate around obstacles or explore potential paths that 
other algorithms might miss due to a more systematic or less exploratory 
approach.

\begin{figure}[ht]
    \centering
    \includesvg[%
        width=\columnwidth,%
        pretex=\relscale{0.85}%
    ]{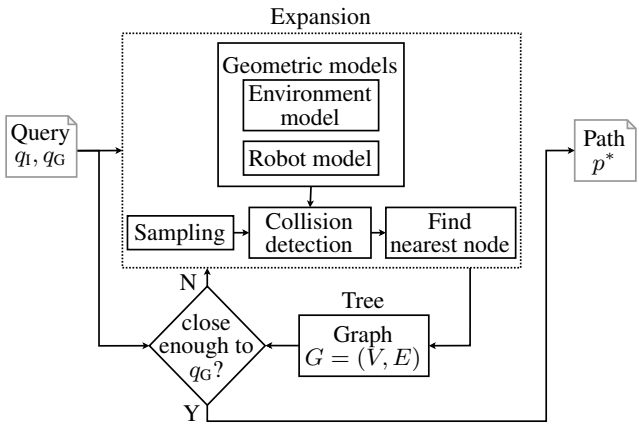}
    \caption{Generic elements of RRT algorithm.}
    \label{fig:rrt_diagram}
\end{figure}

The planning components used in the RRT algorithm are shown in Figure 
\ref{fig:rrt_diagram}. The tree expansion process entails sampling the 
C-space to get a new node, performing collision checking to ensure it is a valid state and then finding the nearest node in the tree (graph) to 
expand. RRT can be considered as a special case of Monte Carlo Tree Search (MCTS) \citep{browne2012survey}, since in the way MCTS uses a \textit{selection} step for choosing the nodes to explore based on some selection policy, RRT also is selecting nodes to \textit{explore} based on its proximity to the newly sampled state. Both algorithms then expand the selected node by adding a new node based on certain rules. Thus, the \textit{heuristic function} and \textit{discrete search} generic components in this context were replaced by finding the nearest node based on the randomly sampled state. This algorithm updates the tree iteratively until a node close enough to the goal is added to the tree.

\subsection{Search-and-Sample Motion Planning}

Search-based motion planning algorithms focus on a discrete representation 
of the state space. They systematically search through a graph or a grid 
that represents possible robot states and paths. Modern search-based motion planning algorithms iteratively construct a graph by expanding nodes (or states) through regular sampling of the continuous configuration space, e.g. with motion primitives \citep{frazzoli2002real}. The search is guided according to a specific strategy, such as depth-first, breadth-first, or best-first search strategies, until 
a path from the start state to the goal state is found. In essence, in 
every iteration a discrete space or graph representation of possible next 
states is obtained by a decomposition method, the most promising next node 
is then selected based on some heuristic, and the next iteration can then 
start from that node, as discussed by \cite{lavalle_planning_2006}.

In this sense, the search and sampling (graph construction) steps are 
interleaved during the planning process, with the validity checks for each 
node performed during sampling. These methods are well-suited for 
environments where a discrete representation can effectively capture the 
necessary details for planning. \cite{cohen_search-based_2010} and \cite{cohen_planning_2011} have proved that these methods guarantee 
optimality with respect to the discretization of the actions and the state 
space, and they are resolution complete, meaning that if a solution exists, it 
will be found given that the resolution of the discretization is sufficiently 
refined.

\begin{figure}[ht]
    \centering
    \includesvg[%
        width=\columnwidth,%
        pretex=\relscale{0.85}%
    ]{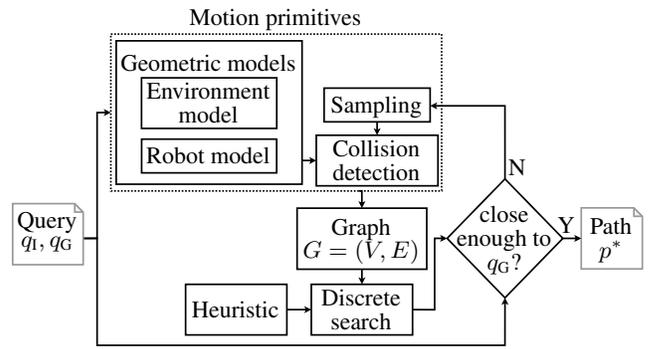}
    \caption{Generic elements of search-and-sample algorithms.}
    \label{fig:smpl_diagram}
\end{figure}

The planning components used in search-and-sample methods are shown in 
Figure \ref{fig:smpl_diagram}. In this method, the action space is sampled 
instead, and a lattice of connected states is obtained. After checking for 
collisions, the valid states are added to the graph which is subsequently 
subjected to a discrete search. This process is repeated iteratively until a state close enough to the goal is sampled and connected to the graph.

\subsection{Trajectory Optimization-based Motion Planning}

Apart from global planning, trajectory optimization is a class of methods that focus on local
refining a given path or trajectory for a robot to minimize or optimize 
a certain objective, such as the path length, energy consumption, or the 
smoothness of the path. It starts with an initial guess of a path and 
iteratively refines this path to find a locally optimal solution that meets 
the constraints and objectives set. This approach is particularly useful in 
manipulator planning, where precise control over the motion is required, 
the dynamics of the system are complex, and the goal is to achieve 
high-quality motions in terms of efficiency, safety, or adherence to 
specific constraints as demonstrated by \cite{kalakrishnan_stomp_2011} and 
\cite{ratliff_chomp_2009}.

While other planning methods, such as sampling-based planners, focus on 
exploring the space to find a feasible path from the start to the goal 
without necessarily optimizing the path, trajectory optimization directly 
aims at improving an existing path to optimize a given objective. While 
sampling-based methods are generally better suited for quickly finding 
a path in high-dimensional spaces or under complex constraints, they do 
not inherently optimize the path. In contrast, 
trajectory optimization takes an existing path and improves it 
according to the specific optimization criteria, making it more suitable 
for applications requiring high-quality motions \citep{mir_survey_2022}. 
Compared to search-based methods, trajectory optimization does not operate 
on a discretized search 
space and it is not constrained by the granularity of the 
discretization. Instead, it is constrained by the computational intensity 
of solving the optimization problem iteratively, which can be more 
scalable to high-dimensional systems.

\begin{figure}[ht]
    \centering
    \includesvg[%
        width=\columnwidth,%
        pretex=\relscale{0.85}%
    ]{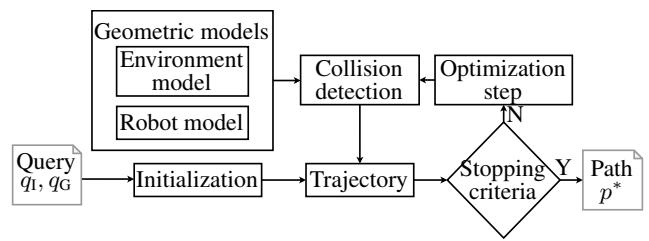}
    \caption{Generic elements of Trajectory Optimization.}
    \label{fig:trajopt_diagram}
\end{figure}

The planning components used in trajectory optimization are shown in 
Figure \ref{fig:trajopt_diagram}. Although this method is fundamentally 
different from the rest, a few parallels can be drawn between its 
functional architecture and the generic case. In particular, instead of a graph, there is a trajectory, and in the place of sampling, there is an initial generation of a sub-optimal trajectory. Accordingly, instead of having a search process, this method iteratively optimizes the initial 
trajectory until all constraints are satisfied within tolerance.

\section{Benchmarking Method}

Properly comparing the two motion planning algorithms’ performance imposes a few requirements. First, we defined a comprehensive set of benchmarks selected to represent a variety of problems the algorithms may encounter in practice. These problems should include different scenarios and task complexities to ensure a broad evaluation spectrum. Second, we decided upon performance metrics that would allow for the quantitative assessment of each algorithm's efficiency, effectiveness, and applicability to different scenarios.

\subsection{Scenarios}

The planning problem instances, or scenarios, we used for benchmarking were identical for both planning frameworks. The following elements in these scenarios are the most critical for our comparison:

\begin{itemize}
    \item \textbf{Robot model}: The specific robot model to be used for 
        planning with a description format compatible with both OMPL and SMPL. 
        In addition, the joints to be used for planning are 
        specified, thus defining the configuration space.
    \item \textbf{Start state}: The start state defines the initial 
        configuration of the robot and is the point to be connected 
        through a viable path to the goal state by the motion planning 
        algorithm.
    \item \textbf{Goal}: The goal state specifies the desired final 
        configuration of the robot and can be a specific point in space or 
        a region specified by allowed ranges for each state variable.
    \item \textbf{Environment}: The environment is a description of the workspace using primitive geometric shapes, encompassing all obstacles 
        and boundaries that the robot must avoid while moving from the 
        start to the goal state.
    \item \textbf{Allowed planning time}: The execution time allowed for 
        the planner to generate a viable path once given the query.
\end{itemize}

\subsection{Evaluation Metrics}

For assessing the performance and effectiveness of the two algorithms, as 
evaluation metrics most useful for this comparison we selected the success 
rate and planning time. Each metric addresses different aspects of 
a planning algorithm's performance and together they offer a comprehensive 
view:

\begin{itemize}
    \item  \textbf{Success Rate}: This metric evaluates the reliability and 
        robustness of a motion planning algorithm. It is defined as the proportion of planning attempts that successfully result in 
        a feasible path from the start to the goal state without 
        collisions. The success rate helps to gauge an algorithm's 
        effectiveness in navigating complex environments and dealing with 
        variations in it.
    \item  \textbf{Planning Time}: This metric measures the efficiency of 
        a motion planning algorithm in terms of computation time. It 
        represents the duration from when the planning request is made 
        until a viable path is found or the algorithm concludes that no 
        such path exists and is crucial in real-time applications, where 
        delays in decision-making could result in operational delays or 
        safety risks. It is desired to minimize planning time while 
        maintaining a high success rate.
\end{itemize}

\section{Implementation and Experiments}

We endeavored to align as best as possible each of the modules in the two 
frameworks aside from the planning algorithms. The intention being to 
enable an objective view of how effectively each planner utilizes the same 
planning modules. This entailed using the same experiment specifications to 
plan with the same robot model description for environments provided in the 
same format.

Therefore, after selecting the benchmarking problem set, we augmented each 
framework according to our needs.

\subsection{Motion Planning frameworks}

The two planning frameworks we compared were the Anytime Replanning A* 
algorithm (ARA*) with motion primitives from the Search-Based Motion Planning Library (SMPL), 
developed by \cite{likhachev_anytime_2005}, and RRT-Connect algorithm with uniform sampling as originally formulated by \cite{kuffner_rrt-connect_2000}, 
implemented in the Open Motion Planning Library (OMPL) and set up using
Robowflex by \cite{kingston_robowflex_2022}. We selected these two algorithms as they are both outstanding examples of the search-based and sampling-based families respectively.

ARA* (Anytime Repairing A*) is a variant of the A* search algorithm that is 
designed to provide solutions quickly and then improve these solutions 
over time, making it suitable for real-time applications. Unlike 
traditional A*, which aims to find the least-cost path before producing any 
solution, ARA* starts by finding a sub-optimal solution quickly and then 
iteratively improves this solution as time permits. It achieves this by 
initially using a relaxed version of the heuristic to speed up the search 
process and then gradually tightening this relaxation to improve the 
solution's optimality. It provides bounds on sub-optimality and improves 
the solution's quality by continuously decreasing the heuristic's inflation 
factor, thereby trading off between search speed and solution optimality. The searched graph of states is constructed using motion primitives, which represent available actions for each robot orientation, as sets of pre-defined and adaptive angles or offsets \citep{cohen_search-based_2010, cohen_planning_2011}.

RRT-Connect is a variant of the RRT algorithm enhancing it by 
simultaneously growing two trees: one from the start state and another from 
the goal state, and then attempting to connect these trees at each 
iteration of the algorithm. This bidirectional approach often finds a path 
more quickly than the unidirectional RRT, especially in high-dimensional 
spaces or environments with complex obstacles. RRT-Connect is efficient in 
finding feasible paths in challenging scenarios, although the paths may not 
be optimal. It's particularly favored for its simplicity and effectiveness 
in various planning domains.

The primary goal of this work was to align all the elements that differ 
between these two implementations in order to obtain a fair 
comparison. More specifically, this entailed using them in the same 
planning scenarios; planning for the same robot; performing collision 
checking in the same manner; and, for ARA*, searching both forward and 
backward, in the same manner that RRT-Connect leverages bidirectional expansion.

\subsection{Scenarios}

The benchmarking planning scenarios used were the same as test sets 
for the Spark, Flame, and RRT-Connect planners by \cite{chamzas_learning_2021}, 
generated with MotionBenchMaker \citep{chamzas_motionbenchmaker_2022}. These are three groups of 100 request 
variations over three environments. All environments contain shelves with 
objects inside them which the robot is requested to either approach or take 
out. The objects are represented by basic cylinder and cube geometric 
shapes.

Each group has a specific kind of variation, among its 100 scenarios, on 
the position and orientation of the shelf and the objects it contains. The 
\texttt{shelf\_zero\_test} group has the shelf in a fixed position while varying 
the locations of the objects in it; the \texttt{shelf\_height\_test} group 
includes additional variation on the shelf height; and in the 
\texttt{shelf\_height\_rot\_test} group the shelf's rotation relative to the 
robot's base reference is additionally varied.

\begin{figure}[ht!]
    \centering
    \begin{subfigure}{0.2\columnwidth}
        \centering
        \includegraphics[width=\columnwidth]{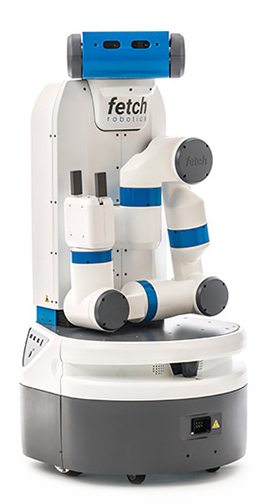}
        \caption{The Fetch robot \citep{wise_fetch_2016}.}
        \label{fig:fetch_robot}
    \end{subfigure}
    \begin{subfigure}{0.39\columnwidth}
        \centering
        \includegraphics[width=\columnwidth]{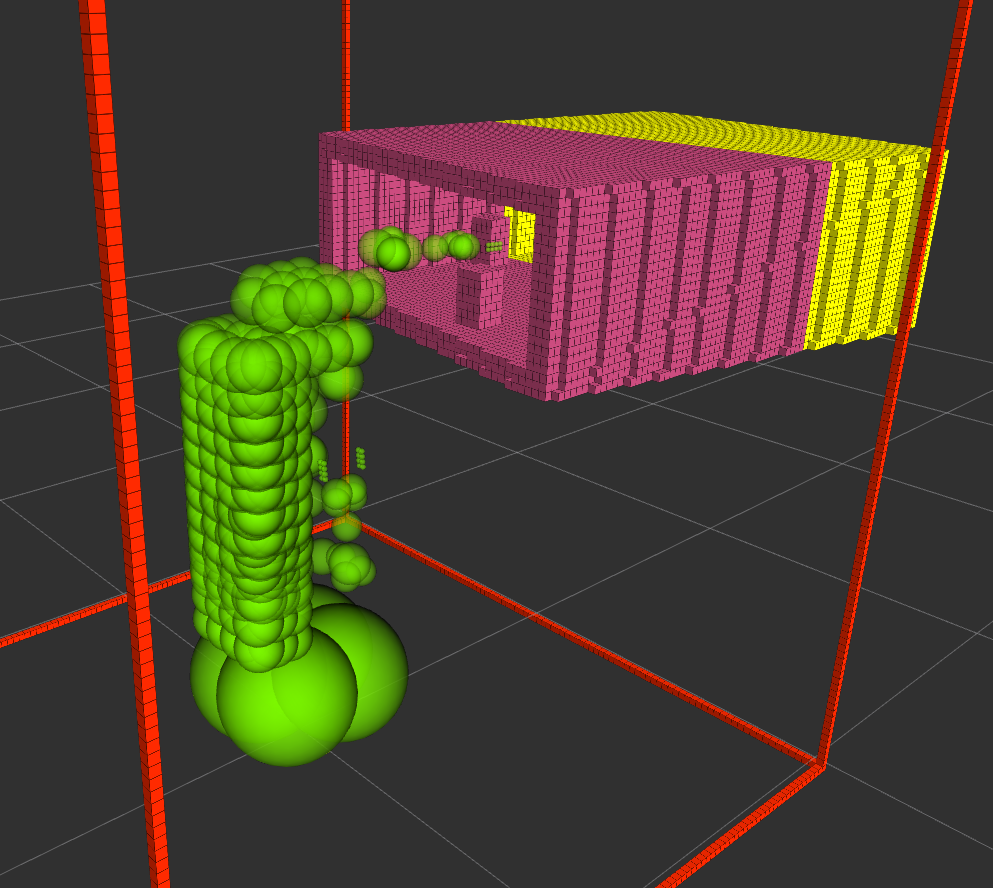}
        \caption{An example of a planning environment.}
        \label{fig:planning_environment}
    \end{subfigure}
    \begin{subfigure}{0.39\columnwidth}
        \centering
        \includegraphics[width=0.77\columnwidth]{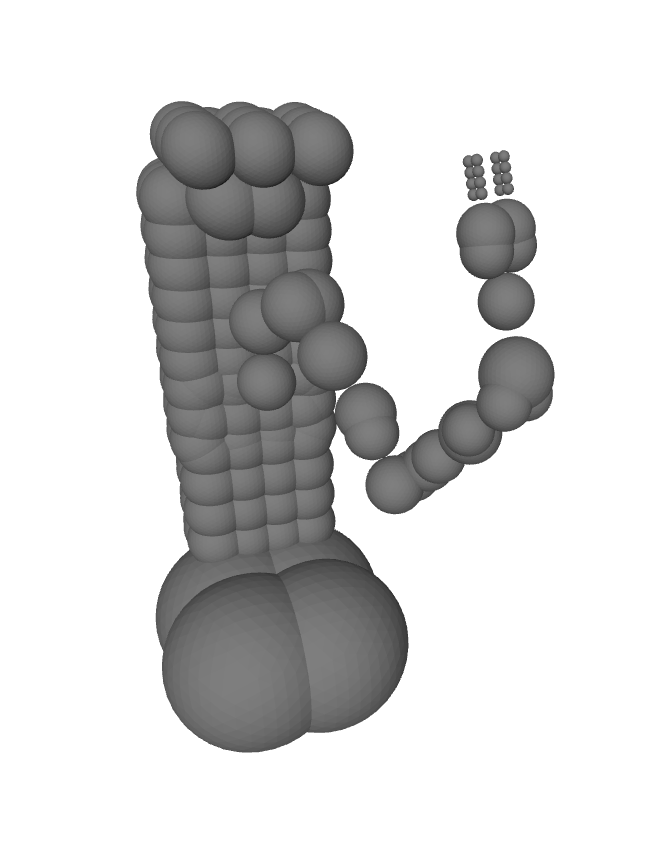}
        \caption{The Fetch robot's spheres-collision model.}
        \label{fig:spheres_model}
    \end{subfigure}
    \caption{The robot and the environment used in benchmarking.}
    \label{fig:geometries}
\end{figure}

The objects in the environments were primitive geometric descriptions of 
standard ROS planning scene messages, in the same format that was used to benchmark 
RRT-Connect with uniform sampling and biased sampling-based planners by 
\cite{chamzas_learning_2021}. An example of an environment as used for 
search-based planning is shown in Figure 
\ref{fig:geometries}\subref{fig:planning_environment}. The start and goal 
were also randomly varied among all scenarios in each group. The Fetch 
robot was used \citep{wise_fetch_2016}, shown in figure \ref{fig:geometries}\subref{fig:fetch_robot}, with the same 
8 planning joints for all scenarios, resulting in an 8-dimensional 
  planning space with 1 prismatic and 
7 rotational degrees of freedom. The allowed planning time was 60'' and was 
the same for every planning request.

\subsection{SMPL}

The first main extension of SMPL involved enabling the use of the 
experiment specification files generated by MotionBenchMaker that are used 
in OMPL. These are two types of YAML formatted files both including 
entries corresponding to standard ROS \textit{MoveIt!} planning message 
types: files containing entries for \texttt{moveit\_msgs/MotionPlanRequest} 
messages, and files containing entries for \texttt{moveit\_msgs/PlanningScene}
messages, as defined by \cite{coleman_reducing_2014}. Originally only 
SMPL-specific definitions for planning environments and requests were 
supported, thus not allowing for automating the benchmarking of ARA* with 
motion primitives on sets of planning problems from other libraries.

The second necessary modification was performing both a forward and 
a backward search with ARA* for each problem. The reasoning behind this 
was to offer search-based planning the same advantage of avoiding becoming 
trapped in convex obstacles in the workspace similar to how RRT-Connect 
bypasses this problem with bidirectional expansion. Backward search has 
been proven by \cite{garrett_backward-forward_2015} to benefit the search 
when planning for manipulators in high-dimensional state spaces. In a case 
where the robot is required to reach inside an opening, search-based 
planning can produce a path for the reverse problem considerably faster 
while that path is at the same time a valid solution to the forward 
problem. In our experiments, each problem was tried using both forward 
and backward search in sequence, and the first solution was kept.

The last major addition was the construction of a Fetch robot description 
for SMPL. While the basic structural description is obtained by the 
corresponding ROS description package, several other configuration files 
were required. The most important are those of the motion primitives 
definitions, the discretization for each joint state, the distance metric 
function that had to include an additional prismatic DoF compared to other 
manipulators, and the collision model of the robot. In this library, the 
robots' collision representations are sets of spheres arranged to resemble 
the actual links' geometries. This collision model is shown in Figure 
\ref{fig:geometries}\subref{fig:spheres_model}.
This method introduces two main challenges for comparing this framework's 
planners to others: first, the collision checking with spheres is considerably faster 
compared to using triangular mesh geometries; and second, 
although the approximation is tailor-made to be as close as possible to the actual geometry, 
it is still prone to inaccuracies compared to the original. To alleviate 
this inconsistency, it is possible to further adjust it.

\subsection{OMPL with Robowflex}

The only modification required on the side of the Robowflex framework 
was performing collision checking in the same way as in SMPL. This was 
achieved by creating a custom Fetch robot URDF description file with 
sphere primitive geometries instead of meshes for each link. In practice, 
every link's mesh geometry was automatically replaced by sphere 
definitions with the same centers and radii as those specified in SMPL's 
model. In this way, \textit{MoveIt!}'s collision checking using primitive geometric 
descriptions was leveraged accelerating the collision checking duration 
phase during RRT-Connect's execution.

\section{Results and Discussion}

\begin{figure}[ht]
    \centering
    \includesvg[width=0.90\columnwidth]{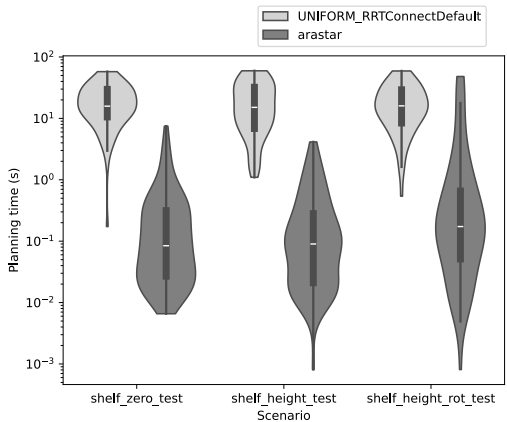}
    \caption{Average planning time for RRT-Connect and ARA* across the 
three different planning scenario groups.}
    \label{fig:planning_time}
\end{figure}

\begin{figure}[ht]
    \centering
    \includesvg[width=0.85\columnwidth]{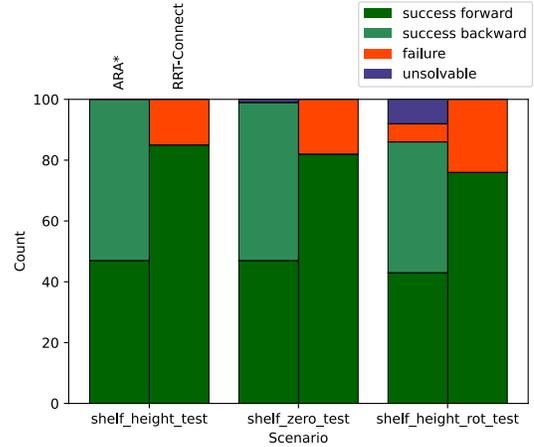}
    \caption{Success rate for RRT-Connect and ARA* across the three different planning scenario groups.}
    \label{fig:success_rate}
\end{figure}

The planning experiments were run on a compute node with an Intel Xeon E5-6248R 24C 3.0GHz and 8 GB of RAM memory per CPU; each experiment executed as a 
single-threaded process.
The distribution of planning times for each planning scenario 
group is shown in Figure \ref{fig:planning_time}. The success rates are 
similarly shown in Figure \ref{fig:success_rate}. The vertical axis for 
the planning time is on a logarithmic scale. The specific numbers of 
successes, failures, and unsolvable cases are presented in Tables 
\ref{tab:results_arastar} and \ref{tab:results_rrt-connect}.

It is worth noting that in total, ARA* failed in 15 out of 
the 300 problems. 9 of these were unsolvable due to collisions in the 
initial or goal state, while for the other 6, it failed to find a path. 
Out of the first 9 unsolvable by ARA*, RRT-Connect succeeded in 6; 
and out of the rest 6 where ARA* failed, RRT-Connect succeeded in 4. 
In total, RRT-Connect failed in 57 out of 300 problems.

\begin{table}[ht!]
\begin{scriptsize}
\caption{Benchmarking results for ARA*.}
\label{tab:results_arastar}
    \begin{tabular}{ r c c c c } 
        Experiment & \makecell{Success \\ forward} & \makecell{Success \\ backward} & failure & unsolvable \\
        \hline
        \texttt{shelf\_zero\_test} & 47 & 53 & 0 & 0 \\
        \texttt{shelf\_height\_test} & 47 & 52 & 0 & 1 \\ 
        \texttt{shelf\_height\_rot\_test} & 43 & 43 & 6 & 8 \\ 
        \hline
    \end{tabular}
\end{scriptsize}
\end{table}

\begin{table}[ht!]
\begin{scriptsize}
\caption{Benchmarking results for RRT-Connect.}
\label{tab:results_rrt-connect}
    \begin{tabular}{ r c c c c } 
        Experiment & \makecell{Success \\ forward} & \makecell{Success \\ backward} & failure & unsolvable \\
        \hline
        \texttt{shelf\_zero\_test} & 85 & - & 15 & 0 \\
        \texttt{shelf\_height\_test} & 82 & - & 18 & 0 \\ 
        \texttt{shelf\_height\_rot\_test} & 76 & - & 24 & 0 \\ 
        \hline
    \end{tabular}
\end{scriptsize}
\end{table}

Based on the performance of the two frameworks in our experiments, we were 
able to make several observations. Compared to ARA*, RRT-Connect exhibits 
more consistent performance in all three groups of problems, both in terms 
of planning time and of success rate. However, despite the more pronounced 
variation, ARA* was approximately 100 times faster in planning and had an 
overall slightly higher success rate.

It is important to note that the motion primitives used for the 
search-based planning of SMPL in this case were appropriately chosen for 
manipulators and were not arbitrary. To achieve similar performance when 
planning for a different system in a different environment it would be 
necessary to configure the motion primitives accordingly. Therefore, the 
superior performance of search-based planning is tied to this type of 
planning problem and is not guaranteed when planning for other robots.

On the other hand, the sampling-based planning required no prior 
adjustments for these particular problems. In principle, it would perform 
equally well when planning for any other system with the same number of 
degrees of freedom. Across the board, sampling-based planning is more 
a more consistent option, while search-based methods can be adjusted to 
perform better in individual cases.

One caveat of this effort is that RRT-Connect's performance was negatively 
impacted by the sphere-collision model. Specifically, mean and variance of 
the planning time were increased, and the success rate was slightly reduced. At 
the same time, owing to the voxelization method in SMPL, some of the 
planning problems were unsolvable by ARA* due to collisions in the initial 
and goal configurations. Although we strove to prepare as accurate a robot 
description and collision model as possible, issues like these are bound to 
occur when trying to bridge the gap between two inherently different 
implementations.

\section{Related work}

The relationship between search-based and sampling-based motion planning 
methods, especially in high-dimensional spaces has been explored by 
\cite{lavalle_relationship_2004}. This work pointed out the advantages of sampling methods 
in such contexts and outlined the importance of discrepancy and dispersion in 
understanding motion planning algorithms' efficacy. However, there has 
been no significant follow-up work to this in the research 
community. \cite{cohen_generic_2012} introduced a more generic infrastructure for 
benchmarking motion planners, comparing notable algorithms like ARA* and 
RRT. However, their benchmarks did not account for adjustments in 
collision models, which could significantly influence the algorithms' 
performance metrics.

Until recently, there was a notable lack of attempts to create 
a comprehensive benchmarking software that extended beyond OMPL by \cite{sucan_open_2012}, as pointed out by \cite{elbanhawi_sampling-based_2014}. 
The introduction of benchmarking capabilities in OMPL highlighted the 
need to establish benchmarking standards that facilitate a thorough 
comparison of sampling-based motion planners. Their work represents 
a significant effort towards creating a common ground for 
benchmarking.

Besides global planning for robot manipulation, \cite{wahab_comparative_2020} compared classical algorithms, like 
Dijkstra's, to meta-heuristic algorithms for mobile robot path planning, 
albeit within the same framework. This approach to benchmarking focused on 
planning modules implemented under one tailor-made framework. Similarly, 
\cite{spahn_local_2022} developed an extensible benchmarking suite for local 
planners which nevertheless requires manual integration of motion planners.

\cite{chamzas_motionbenchmaker_2022} and \cite{orthey_sampling-based_2023} both contributed to 
the expanding landscape of motion planning benchmarks, with the former 
providing an extensible tool for dataset generation and comparison of 
algorithms within OMPL, and the latter comparing several sampling-based 
motion planning methods using OMPL, among other tools. However, search-based motion planning is not treated. Lastly, for the 
extended problem of Task and Motion Planning, \cite{lagriffoul_platform-independent_2018} highlight the challenges in benchmarking, 
pointing out the absence of a common set of metrics, formats, and problems 
accepted by the community. Their proposal for a set of benchmark problems 
and a planner-independent specification format for TAMP challenges seeks to 
unify the community around standard benchmarks.

\section{Conclusion}

In this study, we compared two fundamentally distinct approaches: the sampling-based RRT-Connect and the search-based ARA* utilizing motion primitives. In principle, these are two approaches that solve the motion planning problem by approaching the discretization of the continuous planning domain in distinct ways. Through an experimental setup that ensured equal planning conditions, we illustrated the strengths and limitations of each planning paradigm.

RRT-Connect showed a more uniform performance across various high-dimensional scenarios, underscoring its adaptability and broad applicability. Conversely, the search-based ARA* demonstrated a potential for superior performance, especially when paired with an appropriately tailored action-space sampling scheme, highlighting the significance of customization in optimizing algorithmic efficiency.

ARA* was found capable plan almost 100 times faster than RRT-Connect on average. Nonetheless, owing to the custom robot collision model there was a small set of problems exclusively unsolvable by ARA*. At the same time, the incorporation of the spheres collision model to RRT-Connect’s collision checking resulted in a comparable worsening of its performance.

The implications of this work extend beyond the comparative performance metrics, offering an understanding of the complexities inherent in unifying the benchmarking conditions of two independently developed planning frameworks. Moreover, we hope to provide a new perspective on the basic components of motion planning algorithms and the implications their functional structure has for their performance. We hope that this study will also contribute to bridging the gap between these communities.

However, there remain several directions for further exploration and refinement of our research. Future investigations could focus on aligning SMPL's collision checking with OMPL and Robowflex's mesh 
collision checking. Furthermore, code profiling would provide valuable 
insight as to which phases of the planning process in each method 
take significant computation time. Lastly, metrics for judging the solutions' optimality, such as trajectory length or duration, would be 
particularly informative for determining the preferable use cases of each method.

For future work, search-based algorithms provide a solid foundation for several advancements. The search process can potentially be accelerated by employing burs \citep{lacevic2016burs} as more principled motion primitives. Additionally, prior search experience can be utilized to learn heuristics for guiding motion planning \cite{ajanovic2023value}, or to learn the distance from the optimal path to prune non-promising branches from the search \citep{bokan2024slope}.

\section*{Acknowledgments}
This work was partially supported by the EU Horizon 2020 research and development programme, grant number 952215 (TAILOR). The authors would like to thank Bakir Lacevic for his insightful comments and valuable feedback on this work.

\bigskip

\bibliography{
    references/0_shared,
    references/1_planning_problem,
    references/2_planning_algorithms,
    references/3_method,
    references/4_implementation,
    references/6_related_work
}

\end{document}